\documentclass[10pt,twocolumn,letterpaper]{article}

\usepackage{iccv}
\usepackage{times}
\usepackage{epsfig}
\usepackage{graphicx}
\usepackage{amsmath}
\usepackage{amssymb}

\usepackage{bm}

\usepackage{multirow}
\usepackage{array}
\newcolumntype{L}[1]{>{\raggedright\let\newline\\\arraybackslash\hspace{0pt}}m{#1}}
\newcolumntype{C}[1]{>{\centering\let\newline\\\arraybackslash\hspace{0pt}}m{#1}}
\newcolumntype{R}[1]{>{\raggedleft\let\newline\\\arraybackslash\hspace{0pt}}m{#1}}

\usepackage{graphicx}
\usepackage[justification=centering, labelformat=simple]{subfig}

\usepackage{enumerate}

\newcommand{\specialcell}[2][c]{%
	\begin{tabular}[#1]{@{}c@{}}#2\end{tabular}}

\usepackage[pagebackref=true,breaklinks=true,letterpaper=true,colorlinks,bookmarks=false]{hyperref}

 \iccvfinalcopy 

\def\iccvPaperID{****} 

\ificcvfinal\fi
\begin{document}

\title{Image Super-Resolution via RL-CSC: When Residual Learning Meets Convolutional Sparse Coding}

\author{Menglei Zhang, 
Zhou Liu, 
Lei Yu\\
School of Electronic and Information, Wuhan University, China\\
{\tt\small \{zmlhome, liuzhou, ly.wd\}@whu.edu.cn}
}

\maketitle

\begin{abstract}
We propose a simple yet effective model for Single Image Super-Resolution (SISR), by combining the merits of Residual Learning and Convolutional Sparse Coding (RL-CSC). Our model is inspired by the Learned Iterative Shrinkage-Threshold Algorithm (LISTA). We extend LISTA to its convolutional version and build the main part of our model by strictly following the convolutional form, which improves the network's interpretability. Specifically, the convolutional sparse codings of input feature maps are learned in a recursive manner, and high-frequency information can be recovered from these CSCs. More importantly, residual learning is applied to alleviate the training difficulty when the network goes deeper. Extensive experiments on benchmark datasets demonstrate the effectiveness of our method. RL-CSC ($30$ layers) outperforms several recent state-of-the-arts, \eg, DRRN ($52$ layers) and MemNet ($80$ layers) in both accuracy and visual qualities. Codes and more results are available at \url{https://github.com/axzml/RL-CSC}.
\end{abstract}

\section{Introduction}

\begin{figure}[t]
	\centering
	\includegraphics[width=\linewidth, keepaspectratio]{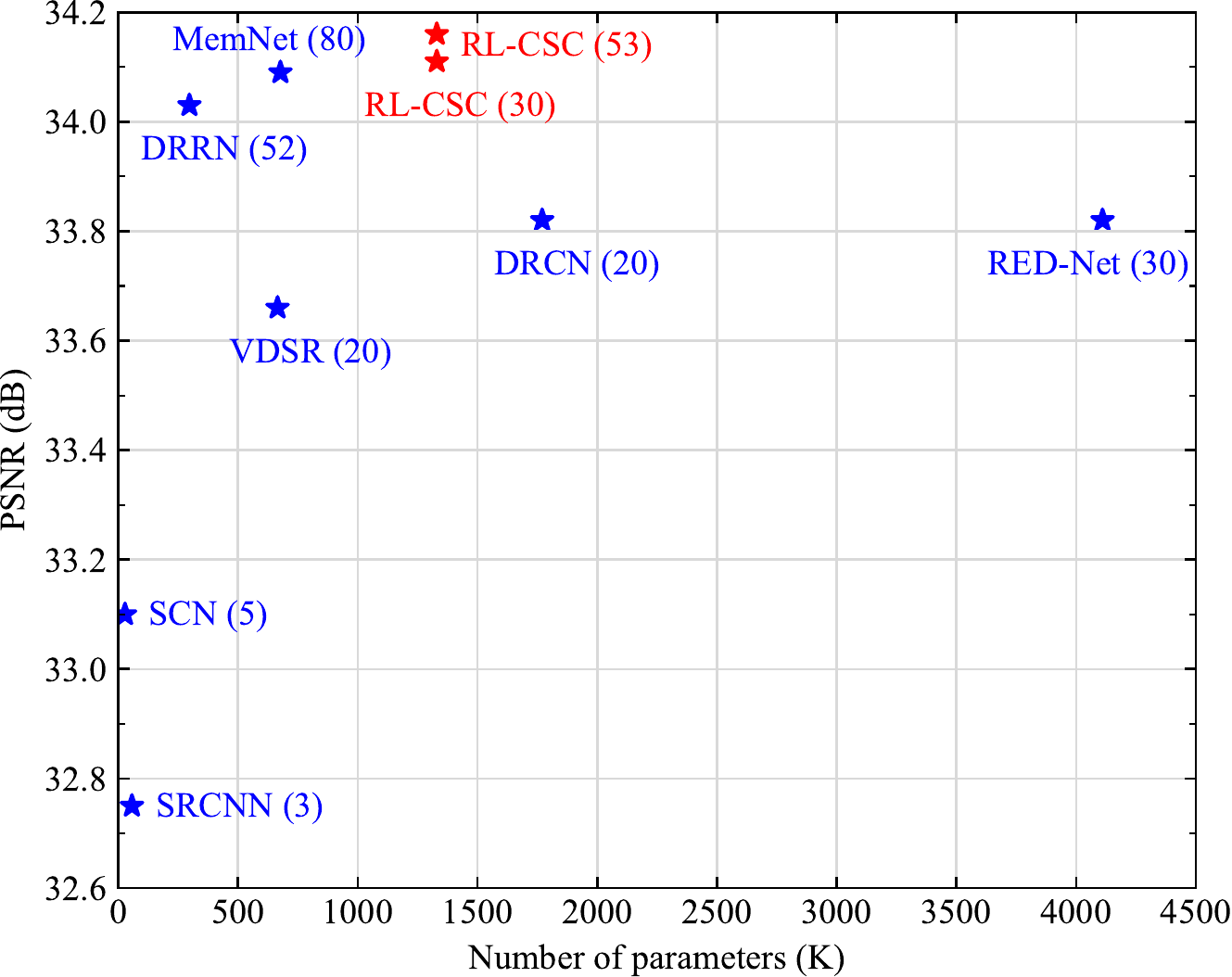}
	\caption{\label{fig:psnr-params} PSNRs of recent CNN models versus the number of parameters for scale factor $\times 3$ on Set5 \cite{Set5:2012}. The number of layers are marked in the parentheses. \textcolor{red}{Red} points represent our models. RL-CSC with $25$ recursions achieves competitive performance with MemNet \cite{MemNet2017}. When increasing the number of recursions without introducing any parameters, the performance of RL-CSC can be further improved.}
\vspace*{-.5cm}
\end{figure}

Single Image Super-Resolution (SISR), which aims to restore a visually pleasing high-resolution (HR) image from its low-resolution (LR) version, is still a challenging task within computer vision research community \cite{NTIRE2017, NTIRE2018}. Since multiple solutions exist for the mapping from LR to HR space, SISR is highly ill-posed and a variety of algorithms, especially the current leading learning-based methods \cite{Aplus2014, SRCNN2016, VDSR2016, DRCN2016, DRRN2017, SRDenseNet2017} are proposed to address this problem.

In recent years, Convolutional Neural Network (CNN) has shown remarkable performance for various computer vision task \cite{ResNet2016, DnCNN2017, Yang:2018wc} owing to its powerful capabilities of learning informative hierarchical representations. Dong \etal~\cite{SRCNN2016} firstly proposed the seminal CNN model for SR termed as SRCNN, which exploits a shallow convolutional neural network to learn a nonlinear LR-HR mapping in an end-to-end manner and dramatically overshadows conventional methods \cite{JianChaoYang2010}. Inspired by VGG-net \cite{VGG2014}, Kim \etal~\cite{VDSR2016} firstly constructed a very deep network up to 20 layers named VDSR, which shows significant improvements over SRCNN. Techniques like skip-connection, adjustable gradient clipping were introduced to mitigate the vanishing-gradient problem when the network goes deeper. Kim \etal further proposed a deeply-recursive convolutional network (DRCN) \cite{DRCN2016} with a very deep recursive layer, and performance can be even improved by increasing recursion depth without new parameters introduced. As the extraordinary success of ResNet \cite{ResNet2016} in image recognition, extensive ResNet or residual units based models for SR have emerged. SRResNet \cite{SRResNet2017} made up of 16 residual units sets up a new state of the art for large upscaling factors ($\times 4$). EDSR \cite{EDSR2017} removes Batch Normalization \cite{BN2015} (BN) layers in residual units and produces astonishing results in both qualitative and quantitative measurements. Tai \etal~\cite{DRRN2017} proposed DRRN in which modified residual units are learned in a recursive manner, leading to a deeper yet concise network. They further introduced memory block to build MemNet \cite{MemNet2017} based on dense connections. In \cite{REDNet2016}, an encoding-decoding network named RED-Net was proposed to take full advantage of many symmetric skip connections.

Despite achieving amazing success in SR, the aforementioned models usually lack convincing analyses about why they worked.
Numerous questions are expected to be explored, \ie, what role each module plays in the network, whether BN is needed, \etc.
 In the past decades, sparse representation with strong theoretical support has been widely used \cite{JianChaoYang2010, Zhang:2015kb} due to its good performance. It’s still valuable even nowadays data-driven models have became more and more popular. Wang \etal~\cite{SCN2015} introduced a sparse coding based network (SCN) for image super-resolution task, by combining the powerful learning ability of neural network and people’s domain expertise of sparse coding,  which fully exploits the approximation of sparse coding learned from LISTA \cite{LISTA2010}. With considerable improvements of SCN over traditional sparse coding methods \cite{JianChaoYang2010} and SRCNN \cite{SRCNN2016} are observed, the authors claim that people’s domain knowledge is still valuable and when it combines with the merits of deep learning, results can benefit a lot. However, layers in SCN are strictly corresponding to each step in the procedure of traditional sparse coding based image SR, so the network still attempts to learn the mapping from LR to HR images. It turns out to be inefficient as indicated in \cite{VDSR2016, DRCN2016, DRRN2017}, which limits the results to be further improved. Moreover, as the experimental results of SCN show no observable advancement when the number of recurrent stages $k$ is increased, the authors finally choose $k = 1$ causing SCN to become a shallow network.

Convolutional Sparse Coding (CSC) has attained much attention from researchers \cite{Zeiler:2010kx, Bristow:2013fw, Heide:2015hq, GarciaCardona:2018id} for years. As CSC inherently takes the consistency constraint of pixels in overlapped patches into consideration, Gu \etal~\cite{CSC-SR2015} proposed CSC based SR (CSC-SR) model and revealed the potential of CSC for image super-resolution over conventional sparse coding methods. In order to build a computationally efficient CSC model, Sreter \etal~\cite{LCSC2018} introduced a convolutional recurrent sparse auto-encoder by extending the LISTA method to a convolutional version, and demonstrated its efficiency in image denoising and inpainting tasks.

To add more interpretability to CNN models for SR and inspire more researches to focus on this topic, we propose a novel model for SR, simple yet effective, to attempt to combine the merits of Residual Learning and Convolutional Sparse Coding (RL-CSC). In a nutshell, the contributions of this paper are three-fold:
\begin{enumerate}
	\itemsep0em 
	\item Unlike many researchers referring to networks proposed in the field of image recognition for inspiration, our model, termed as RL-CSC, is deduced from LISTA. So we provide a new effective way to facilitate model construction, in which every module has well-defined interpretability.
	\item Analyses about the advantages over \cite{SCN2015, DRCN2016, DRRN2017} are discussed in detail.
	\item Thanks to the guidelines of sparse coding theory, RL-CSC ($30$ layers) has achieved competitive results with DRRN \cite{DRRN2017} ($52$ layers) and MemNet \cite{MemNet2017} (up to $80$ layers) in image super-resolution task.  Figure~\ref{fig:psnr-params} shows the performances of several recent CNN models \cite{SRCNN2016, SCN2015, VDSR2016, DRCN2016, REDNet2016, DRRN2017, MemNet2017} in SR task.
\end{enumerate} 

\section{Related Work}
\begin{figure*}
\centering
\includegraphics[width=\textwidth, keepaspectratio] {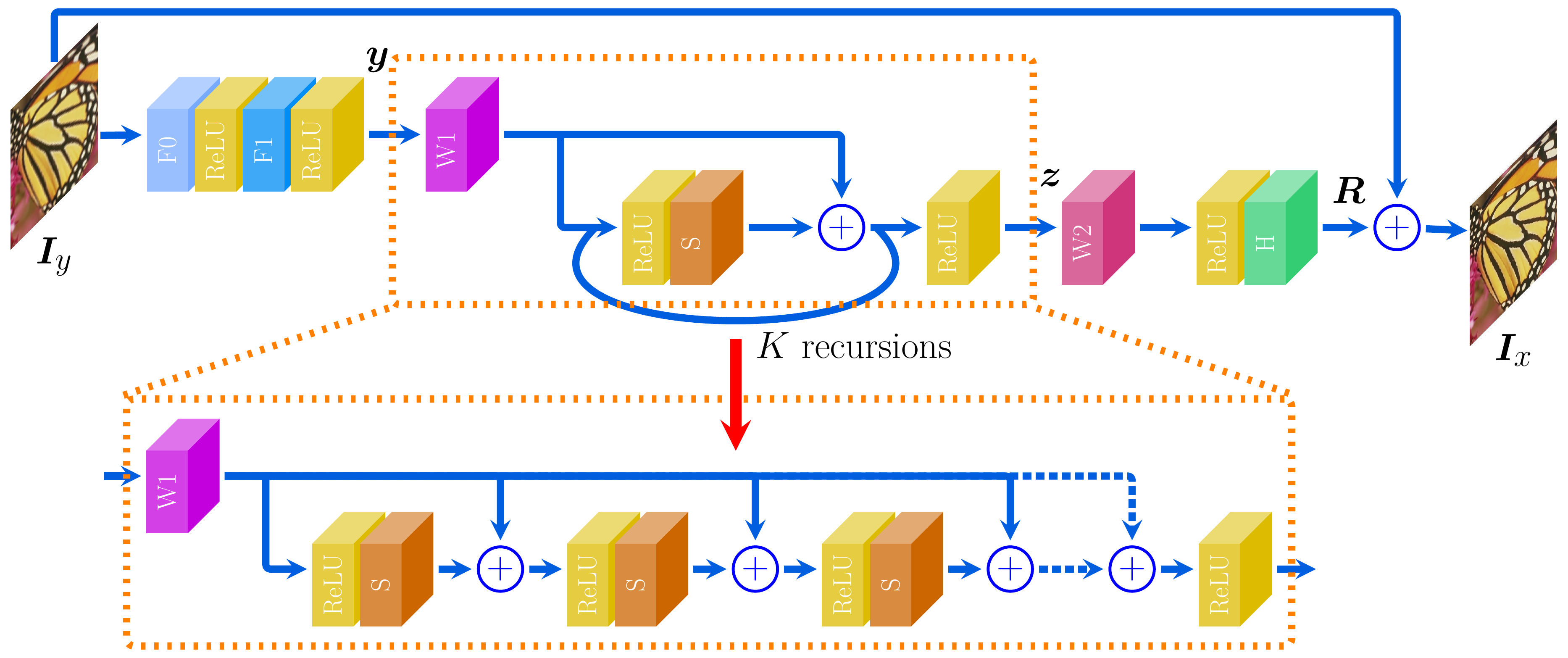}
\caption{\label{fig:model} The proposed RL-CSC framework. Our model takes an interpolated LR image $\bm{I}_y$ as input and predicts the residual component $\bm{R}$. Two convolution layers $\bm{F}_0$ and $\bm{F}_1$ are used for feature extraction and output the feature maps $\bm{y}$, which then go through a convolutional LISTA based sub-network (with $K$ recursions surrounded by the dashed box). When the sparse feature maps $\bm{z}$ are obtained, $\bm{W}_2$ is utilized to recover the features of high-frequency information and the convolution layer $\bm{H}$ maps the features to residual image $\bm{R}$. The final HR image $\bm{I}_x$ can be restored by the addition of ILR image $\bm{I}_y$ and residual image $\bm{R}$. The unfolded version of the recusive sub-network is shown in the bottom.}
\vspace*{-.3cm}
\end{figure*}

\subsection{Sparse Coding and LISTA}

Sparse Coding (SC) has been widely used in a variety of applications such as image classification, super resolution and visual tracking \cite{Zhang:2015kb}. The most popular form of sparse coding attempts to find the optimal sparse code that minimizes the objective function (1), which combines a data fitting term and an  $\ell_1$-norm sparsity-inducing regularization:
\begin{equation}\label{equ:sparse}
	\arg \min _ { \bm { z } } \frac { 1 } { 2 } \| \bm { y } - \bm { D } \bm { z } \| _ { 2 } ^ { 2 } + \lambda \| \bm { z } \| _ { 1 },
\end{equation}
where $\bm{z}\in\mathbb{R}^m$ is the sparse representation of  a given input signal $\bm{y}\in\mathbb{R}^n$ \wrt an $n\times m$ dictionary $\bm{D}$, and regularization coefficient $\lambda$ is used to control the sparsity penalty. $m > n$ is satisfied when $\bm{D}$ is overcomplete.

One popular method to optimize \eqref{equ:sparse} is the so-called Iterative Shrinkage Thresholding Algorithm (ISTA) \cite{ISTA2004, Zhang:2015kb}. At the $k^{\text{th}}$ iteration, the sparse code is updated as:
\begin{equation}\label{equ:ista}
\bm { z } _ { k + 1 } = h _ { \lambda / L } \left( \bm { z } _ { k } + \frac { 1 } { L } \bm { D } ^ { T } ( \bm { y } - \bm { D }\bm { z } _ { k } ) \right),
\end{equation}
where $L \leq\mu_{max}$, and $\mu_{max}$ denotes the largest eigenvalue of $\bm{D}^T\bm{D}$. $h _ { \theta }\left(\cdot\right)$ is an element-wise soft shreshold operator defined as
\begin{equation}\label{equ:soft}
h _ { \theta } ( \alpha ) = \operatorname { sign } ( \alpha ) \max ( | \alpha | - \theta , 0 ).
\end{equation}
However, ISTA suffers from slow convergence speed, which limits its application in real-time situations. To address this issue, Gregor and LeCun~\cite{LISTA2010} proposed a fast algorithm termed as Learned ISTA (LISTA) that produces approximate estimates of sparse code with the power of neural network. LISTA can be obtained by rewriting \eqref{equ:ista} as
\begin{equation}\label{equ:lista}
	\bm { z } _ { \bm { k } + \bm { 1 } } = h _ { \bm{\theta} } \left( \bm{W}_e \bm { y } + \bm{G} \bm { z } _ { \bm { k } } \right),
\end{equation}
where $\frac { 1 } { L } \bm { D } ^ { T }$ is replaced with $\bm{W}_e\in\mathbb{R}^{m\times n}$,  $\bm{I} - \frac { 1 } { L } \bm { D } ^ { T }\bm{D}$ with $\bm{G}\in\mathbb{R}^{m\times m}$ and $\frac{\lambda}{L}$ with a vector $\bm{\theta}\in\mathbb{R}^{m}_{+}$ (so every entry has its own threshold value). Unlike ISTA, parameters $\bm{W}_e$, $\bm{G}$ and $\bm{\theta}$ in LISTA are all learned from training samples using back-propagation procedure. After a fixed number of iterations, the best possible approximation of the sparse code will be produced.

\subsection{Convolutional Sparse Coding (CSC)}

Most of conventional sparse coding based algorithms divide the whole image into overlapped patches and cope with them separately and the consistency constraint, \ie, pixels in the overlapping area of adjacent patches should be exactly the same , is not considered. The convolutional sparse coding (CSC) model \cite{Zeiler:2010kx, Bristow:2013fw, Heide:2015hq,CSC-SR2015,LCSC2018,GarciaCardona:2018id} is inherently suitable for this issue, as it processes the whole image directly:
\begin{equation}\label{equ:csc}
	\arg\min _ { \bm{f}, \bm{Z} } \frac {1}{2} \left\| \bm { Y } - \sum _ { i = 1 } ^ { N } \bm {f} _ { i } \otimes \bm { Z } _ { i } \right\| _ { 2 } ^ { 2 } + \lambda \sum _ { i = 1 } ^ { N } \left\| \bm { Z } _ { i } \right\| _ { 1 },
\end{equation}
where $\bm{Y}\in\mathbb{R}^{m\times n}$ represents an input image, $\{\bm{f}_i\}_{i = 1}^{N}$ is a group of $s\times s$ convolution filters with their respective sparse feature maps $\bm{Z}_i\in\mathbb{R}^{m\times n}$. The reconstruction image can be derived by a summation of convolution results:
\begin{equation}\label{equ:recovery}
\widehat{\bm{Y}} = \sum _ { i = 1 } ^ { N } \bm { f } _ { i } \otimes \bm { Z } _ { i }.
\end{equation}
The current leading strategies on CSC are based on the Alternating Direction Method of Multipliers (ADMM) \cite{Bristow:2013fw,Wohlberg:2014iy, Heide:2015hq,GarciaCardona:2018id}. However, when these methods are utilized to solve \eqref{equ:csc}, the whole training set is optimized at once, tends to cause a heavy memory burden. 

Gu \etal~\cite{CSC-SR2015} proposed a CSC based SR (CSC-SR) method which takes consistency constraint of neighboring patches into consideration for better image reconstruction. SA-ADMM \cite{SA-ADMM2014} is used in their work to alleviate the memory burden issue of ADMM. 

\subsection{Residual Learning}

Residual learning for SR was first introduced in VDSR \cite{VDSR2016} to tackle the vanishing/exploding gradients issue when the network goes deeper. As LR image and HR image are similar to a large extent, fitting the residual mapping seems easier for optimization. Given a training set of $N$ LR-HR pairs $\{ \bm{I}_y^{(i)}, \bm{I}_x^{(i)} \}_{i=1}^{N}$, the residual image is defined as $\bm{r}^{(i)} = \bm{I}_x^{(i)} - \bm{I}_y^{(i)}$, the goal is to learn a model $f$ with parameters $\bm{\Theta}$ that minimizes the following objective function:
\begin{equation}\label{equ:residual}
\mathcal{L}(\bm{\Theta}) = \frac{1}{N}\sum_{i = 1}^{N}\left\Vert f\left(\bm{I}_y^{(i)}\right) - \bm{r}^{(i)}\right\Vert^2_2.
\end{equation}

VDSR \cite{DRCN2016} uses a single skip connection to link the input Interpolated LR (ILR) image and the final output of the network for HR image reconstruction, termed as \textit{Global Residual Learning} (GRL), which benefits both the convergency speed and the reconstruction accuracy a great deal. In DRRN \cite{DRRN2017}, both \textit{Global} and \textit{Local Residual Learning} (LRL) are adopted to help the gradient flow. 

\section{Proposed Method}
\subsection{Feature Extraction from ILR Image}

As illustrated in Figure~\ref{fig:model}, our model takes the Interpolated Low-Resolution (ILR) image $\bm{I}_y$ as input, and predicts the output HR image $\bm{I}_x$. Two convolution layers, $\bm{F}_0\in\mathbb{R}^{n\times c\times s\times s}$ consisting of $n$ filters of spatial size $c\times s\times s$ and $\bm{F}_1\in\mathbb{R}^{n\times n\times s\times s}$ containing $n$ filters of spatial size $n\times s\times s$ are utilized for hierarchical features extraction from ILR image:
\begin{equation}\label{equ:extract}
\bm{y} = ReLU\Big(\bm{F}_1\otimes ReLU(\bm{F}_0\otimes\bm{I}_y)\Big),
\end{equation}
where $\otimes$ is the convolution operator, and $ReLU(\cdot)$ denotes the Rectified Linear Unit (ReLU) activation function.

\subsection{Learning CSC of ILR Features}
\label{subsec:csc}

It’s worth noting that CSC model can be considered as a special case of conventional SC model, for the convolution operation can be viewed as a matrix multiplication by converting one of the inputs into a Toeplitz matrix. So the CSC model \eqref{equ:csc} has a similar structure to the traditional SC model \eqref{equ:sparse} when the convolution operation is transformed to matrix multiplication. In addition, LISTA is an efficient and effective tool to learn the approximate sparse coding vector of \eqref{equ:sparse}. It takes the exact form of equation \eqref{equ:lista} with the weights $\bm{W}_e$ and $\bm{G}$ represented as linear layers and  it can be viewed as a feed-forward neural network with $\bm{G}$ shared over layers.

In order to efficiently solving \eqref{equ:csc}, we extend \eqref{equ:lista} to its convolutional version by substituting $\bm{W}_e\in\mathbb{R}^{m\times n}$ for $\bm{W}_1\in\mathbb{R}^{m\times n\times s\times s}$, $\bm{G}\in\mathbb{R}^{m\times m}$ for $\bm{S}\in\mathbb{R}^{m\times m\times s\times s}$. The convolutional case of \eqref{equ:lista} can be reformulated as:
\begin{equation}\label{equ:convlista}
\bm{z} _ { k + 1 } = h _ { \bm { \theta } } \left( \bm { W } _ { 1 } \otimes\bm { y } + \bm { S } \otimes\bm { z } _ { k } \right).
\end{equation}
The sparse feature maps $\bm{z}\in\mathbb{R}^{m\times m\times c\times c}$ are learned after $K$ recursions.

As for the activation function $h_{\bm{\theta}}$, \cite{papyan2017convolutional} reveals two important facts: (1) the expressiveness of the sparsity inspired model is not affected even by restricting the coefficients to be nonnegative; (2) the ReLU and the soft nonnegative thresholding operator are equal, that is:
\begin{equation}\label{equ:relu}
h_{\bm{\theta}}^{+}(\bm{\alpha}) = \max(\bm{\alpha} - \bm{\theta}, 0) = ReLU(\bm{\alpha} - \bm{\theta}).
\end{equation}
So we choose ReLU as activation function in RL-CSC.

\subsection{Recovery of Residual Image}

When the sparse feature maps $\bm{z}$ are obtained, they're then fed into a convolution layer $\bm{W}_2\in\mathbb{R}^{m\times n\times s\times s}$ to recover the features of high-frequency information. The last convolution layer $\bm{H}\in\mathbb{R}^{c\times n\times s\times s}$ is used for high-frequency information reconstruction:
\begin{equation}\label{equ:residual-result}
\bm{R} = \bm{H}\otimes ReLU(\bm{W}_2 \otimes \bm{z} ).
\end{equation}
Note that we pad zeros before all convolution operations to keep all the feature maps have the same size, which is a common strategy used in a variety of methods \cite{VDSR2016, DRCN2016, DRRN2017, MemNet2017}. So the residual image $\bm{R}$ has the same size as the input ILR image $\bm{I}_y$, and the final HR image $\bm{I}_x$ will be reconstructed by the addition of $\bm{I}_y$ and $\bm{R}$:
\begin{equation}\label{equ:add}
\bm{I}_x = \bm{I}_y + \bm{R}.
\end{equation}
We build our model by strictly following these analyses.

\subsection{Network Structure}

The entire network structure of RL-CSC is illustrated in Figure~\ref{fig:model}. There are totally 6 trainable layers in our model: two convolution layers $\bm{F}_0$ and $\bm{F}_1$ used for feature extraction, $\bm{W}_1$ and $\bm{S}$ for learning CSC,  $\bm{W}_2$ and $\bm{H}$ for residual image reconstruction. The weight of $\bm{S}$ is shared during every recursion. When $K$ recursions are applied in the training process, the depth $d$ of the network can be calculated as:
\begin{equation}\label{equ:depth}
d = K + 5.
\end{equation}

The loss function of Mean Square Error (MSE) is exploited in our training process. Given $N$ LR-HR image patch pairs  $\{ \bm{I}_y^{(i)}, \bm{I}_x^{(i)} \}_{i=1}^{N}$ as a training set, our goal is to minimize the following objective function with RL-CSC:
\begin{equation}\label{equ:loss}
\mathcal{L}(\bm{\Theta}) = \frac{1}{N}\sum_{i = 1}^{N}\left\Vert \text{RL-CSC}\left(\bm{I}_y^{(i)}\right) + \bm{I}_y^{(i)} - \bm{I}_x^{(i)}\right\Vert^2_2,
\end{equation}
where $\bm{\Theta}$ denotes the learnable parameters. Stochastic gradient descent (SGD) is used for optimization and we implement our model using the PyTorch \cite{pytorch2017} framework.

\section{Discussions}
\begin{figure}
\centering
\renewcommand{\thesubfigure}{(a)}
\subfloat[DRRN]{\includegraphics[width=.24\linewidth, keepaspectratio]{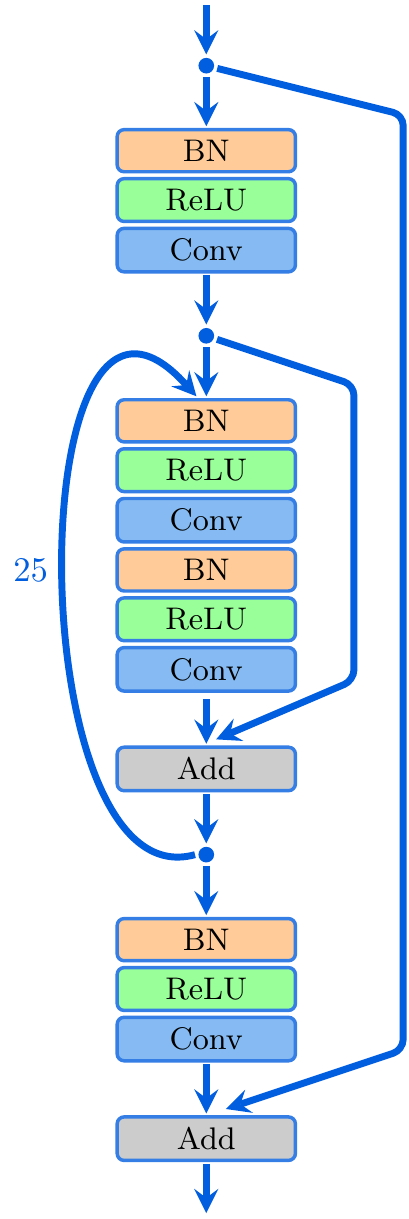}}
\renewcommand{\thesubfigure}{(b)}
\subfloat[SCN]{\includegraphics[width=.25\linewidth, keepaspectratio]{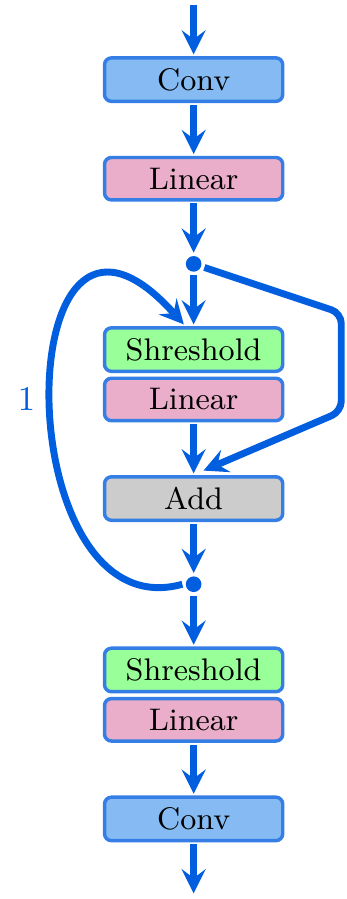}}
\renewcommand{\thesubfigure}{(c)}
\subfloat[DRCN]{\label{subfig:drcn}\includegraphics[width=.233\linewidth, keepaspectratio]{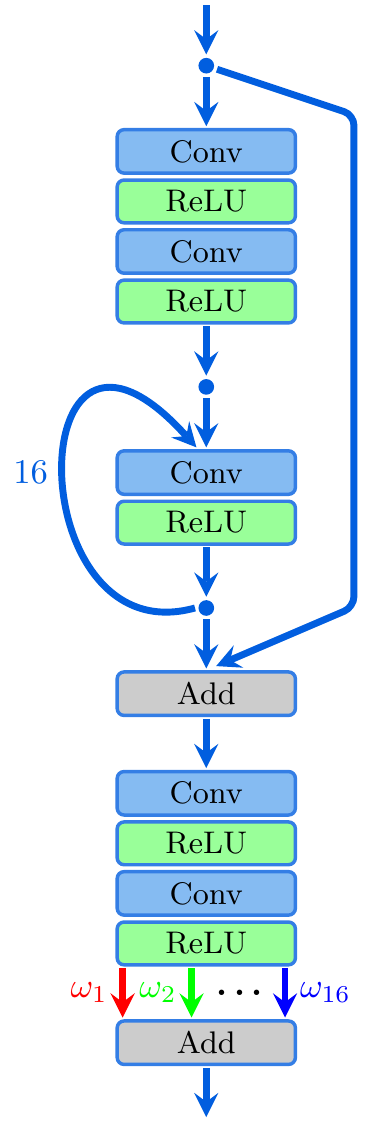}}
\renewcommand{\thesubfigure}{(d)}
\subfloat[RL-CSC]{\includegraphics[width=.23\linewidth, keepaspectratio]{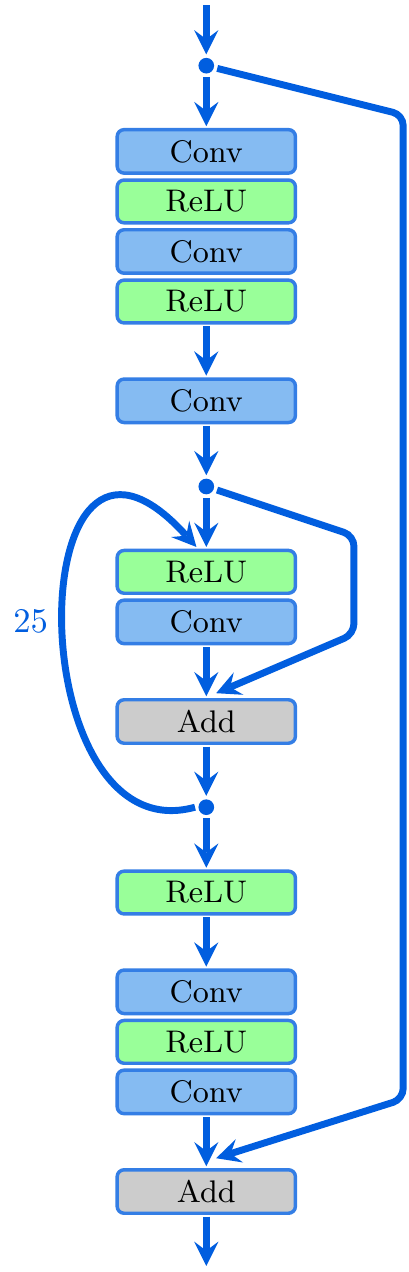}}
\caption{\label{fig:comparisons} Network structures of: (a) DRRN \cite{DRRN2017}. (b) SCN \cite{SCN2015}. (c) DRCN \cite{DRCN2016}. (d) Our model.}
\vspace*{-.5cm}
\end{figure}

In this section, we discuss the advantages of RL-CSC over several recent CNN models for SR with recursive learning strategy applied. Specifically, DRRN \cite{DRRN2017}, SCN \cite{SCN2015} and DRCN \cite{DRCN2016} are used for comparison. The simplified structures of these models are shown in Figure~\ref{fig:comparisons}.   ``Conv’’ is the abbreviation for Convolution layer, ``BN’’ represents Batch Normalization \cite{BN2015}, ``Linear’’ stands for Linear layer and ``Shreshold’’ means Soft Shreshold operator. The digits on the left of the recursion line is the number of recursions.

\textbf{Difference to DRRN}. The main part of DRRN \cite{DRRN2017} is the recursive block structure, in which several residual units are stacked. To further improve the performance, a multi-path structure (all residual units share the same input) and a pre-activation structure (activation layers come before the weight layers) are utilized. These strategies are proved to be effective. The interesting part is, RL-CSC, deduced from LISTA \cite{LISTA2010}, includes a multi-path structure and uses pre-activation inherently. In addition, guided by \eqref{equ:convlista}, RL-CSC contains no BN layers at all. BN consumes much amount of GPU memory and increases computational complexity. Experiments on this topic are conducted in Section~\ref{subsec:recursive}. Furthermore, every module in RL-CSC has a good interpretability, which helps the choice of parameter settings for better performance. Experimental results on benchmark datasets under commonly-used assessments demonstrate the superiority of RL-CSC in Section~\ref{subsec:start-of-the-arts}.

\textbf{Difference to SCN}. There are three main differences between SCN \cite{SCN2015} and RL-CSC. Firstly, RL-CSC ($30$ layers) is much deeper than SCN ($5$ layers). As indicated in \cite{VDSR2016}, a deeper network will have a larger receptive filed, which means the network can utilize more contextual information in an image to infer image details. Secondly, we extend LISTA to its convolutional version in \eqref{equ:convlista}, instead of using linear layers, so more hierarchical information will be extracted. Last but not the least, RL-CSC adopts residual learning, which is a powerful tool for training deeper networks. With the help of residual learning, we can use more recursions, \ie, $25$, even $48$, to achieve better performance.

\textbf{Difference to DRCN}. In the recursive part, RL-CSC differs with DRCN \cite{DRCN2016} in two aspects. One for Local Residual Learning \cite{DRRN2017} (LRL) and the other is pre-activation. Besides, DRCN is not easy to train, so recursive-supervision and skip-connection are introduced to facilitate network to converge. Moreover, an ensemble strategy (In Figure~\ref{subfig:drcn}, the final output is the weighted average of all intermediate predictions) is used to further improve the performance. RL-CSC is relived from these strategies and can be easily trained with more recursions. Advantages of RL-CSC are further illustrated in Section~\ref{subsec:start-of-the-arts}.

{
\begin{table*}
\footnotesize
\centering
\begin{tabular}{|C{1.2cm}|C{.9cm}|C{1.6cm}|C{1.6cm}|C{1.6cm}|C{1.6cm}|C{1.6cm}|C{1.6cm}||C{1.6cm}|}
	\hline
	Dataset & Scale & Bicubic & SRCNN & VDSR & DRCN & DRRN & MemNet & RL-CSC \\\hline\hline
	\multirow{3}{*}{Set5} & $\times2$ & $33.66$/$0.9299$ & $36.66$/$0.9542$ & $37.53$/$0.9587$ & $37.63$/$0.9588$ & $37.74$/$0.9591$ & $\textcolor{blue}{37.78}$/$\textcolor{blue}{0.9597}$ & $\textcolor{red}{37.79}$/$\textcolor{red}{0.9600}$ \\
	& $\times3$ & $30.39$/$0.8682$ & $32.75$/$0.9090$ & $33.66$/$0.9213$ & $33.82$/$0.9226$ & $34.03$/$0.9244$ & $\textcolor{blue}{34.09}$/$\textcolor{blue}{0.9248}$ & $\textcolor{red}{34.11}$/$\textcolor{red}{0.9254}$ \\
	& $\times4$ & $28.42$/$0.8104$ & $30.48$/$0.8628$ & $31.35$/$0.8838$ & $31.53$/$0.8854$ & $31.68$/$0.8888$ & $\textcolor{blue}{31.74}$/$\textcolor{blue}{0.8893}$ & $\textcolor{red}{31.82}$/$\textcolor{red}{0.8907}$\\\hline\hline
	\multirow{3}{*}{Set14} & $\times2$ & $30.24$/$0.8688$ & $32.45$/$0.9067$ & $33.03$/$0.9124$ & $33.04$/$0.9118$ & $33.23$/$0.9136$ & $\textcolor{blue}{33.28}$/$\textcolor{blue}{0.9142}$ & $\textcolor{red}{33.33}$/$\textcolor{red}{0.9152}$ \\
	& $\times3$ & $27.55$/$0.7742$ & $29.30$/$0.8215$ & $29.77$/$0.8314$ & $29.76$/$0.8311$ & $29.96$/$0.8349$ & $\textcolor{red}{30.00}$/$\textcolor{blue}{0.8350}$ & $\textcolor{blue}{29.99}$/$\textcolor{red}{0.8359}$ \\
	& $\times4$ & $26.00$/$0.7027$ & $27.50$/$0.7513$ & $28.01$/$0.7674$ & $28.02$/$0.7670$ & $28.21$/$0.7721$ & $\textcolor{blue}{28.26}$/$\textcolor{blue}{0.7723}$ & $\textcolor{red}{28.29}$/$\textcolor{red}{0.7741}$\\\hline\hline
	\multirow{3}{*}{BSD100} & $\times2$ & $29.56$/$0.8431$ & $31.36$/$0.8879$ & $31.90$/$0.8960$ & $31.85$/$0.8942$ & $32.05$/$0.8973$ & $\textcolor{blue}{32.08}$/$\textcolor{blue}{0.8978}$ & $\textcolor{red}{32.09}$/$\textcolor{red}{0.8985}$ \\
	& $\times3$ & $27.21$/$0.7385$ & $28.41$/$0.7863$ & $28.82$/$0.7976$ & $28.80$/$0.7963$ & $28.95$/$\textcolor{blue}{0.8004}$ & $\textcolor{blue}{28.96}$/$0.8001$ & $\textcolor{red}{28.99}$/$\textcolor{red}{0.8021}$ \\
	& $\times4$ & $25.96$/$0.6675$ & $26.90$/$0.7101$ & $27.29$/$0.7251$ & $27.23$/$0.7233$ & $27.38$/$\textcolor{blue}{0.7284}$ & $\textcolor{blue}{27.40}$/$0.7281$ & $\textcolor{red}{27.44}$/$\textcolor{red}{0.7302}$\\\hline\hline
	\multirow{3}{*}{Urban100} & $\times2$ & $26.88$/$0.8403$ & $29.50$/$0.8946$ & $30.76$/$0.9140$ & $30.75$/$0.9133$ & $31.23$/$0.9188$ & $\textcolor{blue}{31.31}$/$\textcolor{blue}{0.9195}$ & $\textcolor{red}{31.36}$/$\textcolor{red}{0.9207}$ \\
	& $\times3$ & $24.46$/$0.7349$ & $26.24$/$0.7989$ & $27.14$/$0.8279$ & $27.15$/$0.8276$ & $27.53$/$\textcolor{blue}{0.8378}$ & $\textcolor{blue}{27.56}$/$0.8376$ & $\textcolor{red}{27.64}$/$\textcolor{red}{0.8403}$ \\
	& $\times4$ & $23.14$/$0.6577$ & $24.52$/$0.7221$ & $25.18$/$0.7524$ & $25.14$/$0.7510$ & $25.44$/$\textcolor{blue}{0.7638}$ & $\textcolor{blue}{25.50}$/$0.7630$ & $\textcolor{red}{25.59}$/$\textcolor{red}{0.7680}$\\\hline
\end{tabular}
\caption{\label{tab:psnr}Benchmark results. Average PSNR/SSIMs for scale factor $\times2$, $\times3$ and $\times4$ on datasets Set5, Set14, BSD100 and Urban100. \textcolor{red}{Red} color indicates the best performance and \textcolor{blue}{blue} color indicates the second best performance.}
\end{table*}
}

{
	\begin{table*}
		\footnotesize
		\centering
		\begin{tabular}{|C{1.2cm}|C{.8cm}|C{1.2cm}|C{1.5cm}|C{1.5cm}|C{1.5cm}|C{1.7cm}|C{1.9cm}|C{1.5cm}|}
			\hline
			Dataset & Scale & Bicubic & SRCNN \cite{SRCNN2016} & SelfEx \cite{Urban100:2015} & VDSR \cite{VDSR2016} & DRRN\_B1U9 & DRRN\_B1U25 & RL-CSC \\\hline\hline
			\multirow{3}{*}{Set5} & $\times2$ & $6.083$ & $8.036$ & $7.811$ & $8.569$ & $8.583$ & $\textcolor{blue}{8.671}$ & $\textcolor{red}{9.095}$ \\
			& $\times3$ & $3.580$ & $4.658$ & $4.748$ & $5.221$ & $5.241$ & $\textcolor{blue}{5.397}$ & $\textcolor{red}{5.565}$ \\
			& $\times4$ & $2.329$ & $2.991$ & $3.166$ & $3.547$ & $3.581$ & $\textcolor{blue}{3.703}$ & $\textcolor{red}{3.791}$\\\hline\hline
			\multirow{3}{*}{Set14} & $\times2$ & $6.105$ & $7.784$ & $7.591$ & $8.178$ & $8.181$ & $\textcolor{blue}{8.320}$ & $\textcolor{red}{8.656}$ \\
			& $\times3$ & $3.473$ & $4.338$ & $4.371$ & $4.730$ & $4.732$ & $\textcolor{blue}{4.878}$ & $\textcolor{red}{4.992}$ \\
			& $\times4$ & $2.237$ & $2.751$ & $2.893$ & $3.133$ & $3.147$ & $\textcolor{blue}{3.252}$ & $\textcolor{red}{3.324}$\\\hline\hline
			\multirow{3}{*}{Urban100} & $\times2$ & $6.245$ & $7.989$ & $7.937$ & $8.645$ & $8.653$ & $\textcolor{blue}{8.917}$ & $\textcolor{red}{9.372}$ \\
			& $\times3$ & $3.620$ & $4.584$ & $4.843$ & $5.194$ & $5.259$ & $\textcolor{blue}{5.456}$ & $\textcolor{red}{5.662}$ \\
			& $\times4$ & $2.361$ & $2.963$ & $3.314$ & $3.496$ & $3.536$ & $\textcolor{blue}{3.676}$ & $\textcolor{red}{3.816}$\\\hline
		\end{tabular}
		
		\caption{\label{tab:ifc}Benchmark results. Average IFCs for scale factor $\times2$, $\times3$ and $\times4$ on datasets Set5, Set14 and Urban100. \textcolor{red}{Red} color indicates the best performance and \textcolor{blue}{blue} color indicates the second best performance.}
		\vspace*{-.4cm}
	\end{table*}
}

\section{Experimental Results}
In this section, performances of our method on four benchmark datasets are evaluated. We first give a brief introduction to the datasets used for training and testing. Then the implementation details are provided. Finally, comparisons with state-of-the-arts are presented and more analyses about RL-CSC are illustrated.

\subsection{Datasets}

By following \cite{VDSR2016, DRRN2017}, our training set consists of $291$ images, where $91$ of these images are from Yang \etal~\cite{JianChaoYang2010} with the addition of $200$ images from Berkeley Segmentation Dataset \cite{BSD100:2001}. During testing, we choose the dataset \textbf{Set5} \cite{Set5:2012}, and \textbf{Set14} \cite{Set14:2010} which are widely used for benchmark. Moreover, the \textbf{BSD100} \cite{BSD100:2001}, consisting of $100$ natural images are used for testing. Finally, the \textbf{Urban100} of $100$ urban images introduced by Huang \etal~\cite{Urban100:2015} is also employed. Both the Peak Signal-to-Noise Ratio (PSNR) and the Structural SIMilarity (SSIM) on Y channel (\ie, luminance) of transformed YCbCr space are calculated for evaluation.

\subsection{Implementation details}
\label{subsec:details}

To enlarge the training set, data augmentation, which includes flipping (horizontally and vertically), rotating ($90$, $180$, and $270$ degrees), scaling ($0.7$, $0.5$ and $0.4$), is performed on each image of $291$-image dataset. In addition, inspired by prior works, \ie, VDSR \cite{VDSR2016} and DRRN \cite{DRRN2017}, we also train a \textit{single} multi-scale model, which means scale augmentation is exploited by combining images of different scales ($\times 2$, $\times 3$ and $\times 4$) into one training set. Not only for the network scalability, but the fair comparison with other state-of-the-arts. Furthermore, all training images are partitioned into $33\times 33$ patches with the stride of $33$, providing a total of $1,929,728$ LR-HR training pairs.

The dimensions of all convolution layers are determined as follows: $\bm{F}_0\in\mathbb{R}^{128\times 1\times 3\times 3}$, $\bm{F}_1\in\mathbb{R}^{128\times 128\times 3\times 3}$, $\bm{W}_1\in\mathbb{R}^{256\times 128 \times 3\times 3}$, $\bm{S}\in\mathbb{R}^{256\times 256\times 3\times 3}$, $\bm{W}_2\in\mathbb{R}^{256\times 128\times 3\times 3}$, $\bm{H}\in\mathbb{R}^{1\times 128\times 3\times 3}$. As for the number of recursion, we choose $K=25$ in our final model so the depth of RL-CSC is $30$ according to \eqref{equ:depth}. Further discussions about the number of network parameters and $K$ will be illustrated in Section \ref{subsec:recursive}

We follow the same strategy as He \etal~\cite{He:2015} for weight initialization where all weights are drawn from a normal distribution with zero mean and variance $2/n_{out}$, where $n_{out}$ is the number of output units. The network is optimized using SGD with mini-batch size of $128$, momentum parameter of $0.9$ and weight decay of $10^{-4}$. The learning rate is initially set to $0.1$ and then decreased by a factor of $10$ every $10$ epochs. We train a total of $35$ epochs as no further improvements of the loss are observed after that. For maximal convergence speed, we utilize the adjustable gradient clipping strategy stated in \cite{VDSR2016}, with gradients clipped to $[-\theta, \theta]$, where $\theta = 0.4$ is the gradient clipping parameter.  A NVIDIA Titan Xp GPU is used to train our model of $K=25$, which  takes approximately four and a half days .

\subsection{Comparison with State of the Arts}
\label{subsec:start-of-the-arts}

We now compare the proposed RL-CSC model with other state-of-the-arts in recent years. Specifically, SRCNN \cite{SRCNN2016}, VDSR \cite{VDSR2016}, DRCN \cite{DRCN2016}, DRRN \cite{DRRN2017} and MemNet \cite{MemNet2017} are used for benchmarks. All of these models apply bicubic interpolation to the original LR images before passing them to the networks. As the prior methods crop image pixels near borders before evaluation, for fair comparison, we crop the pixels to the same amount as well, even if this is unnecessary for our method.

Table~\ref{tab:psnr} shows the PSNR and SSIM on the four benchmark testing sets, and results of other methods are obtained from \cite{VDSR2016, DRCN2016, DRRN2017, MemNet2017}. Our model RL-CSC with $30$ layers outperforms DRRN ($52$ layers) and MemNet ($80$ layers) in all datasets and scale factors (both PSNR and SSIM).

Furthermore, the metric Information Fidelity Criterion (IFC), which has the highest correlation with perceptual scores for SR evaluation \cite{IFC2014},  is also used for comparison. Experimental results are summarized in Table~\ref{tab:ifc}. The IFCs of \cite{SRCNN2016, Urban100:2015,  VDSR2016} and DRRN are obtained from \cite{DRRN2017}. By following \cite{DRRN2017}, BSD100 is not evaluated. It’s obvious that our method achieves better performances than other methods in all datasets and scale factors.

\begin{figure*}
\renewcommand{\thesubfigure}{} 
\vspace*{-.2cm}
\subfloat[Ground Truth \protect\\ (PSNR, SSIM)]{\label{sub:GT:8023}\includegraphics[%
	width=0.125\textwidth, keepaspectratio]{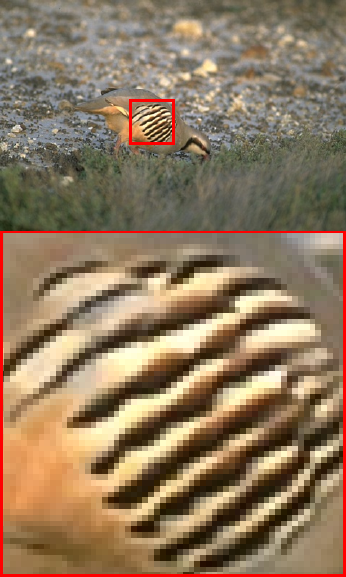}}
\subfloat[Bicubic\protect\\ ($28.50$, $0.8285$)]{\label{sub:Bicubic:8023}\includegraphics[%
	width=0.125\textwidth, keepaspectratio]{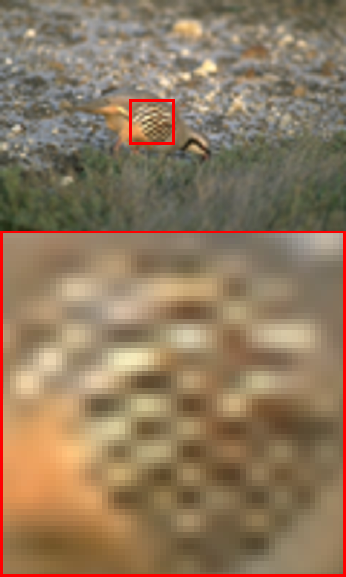}}
\subfloat[SRCNN \cite{SRCNN2016} \protect\\ ($29.40$, $0.8561$)]{\label{sub:SRCNN:8023}\includegraphics[%
	width=0.125\textwidth, keepaspectratio]{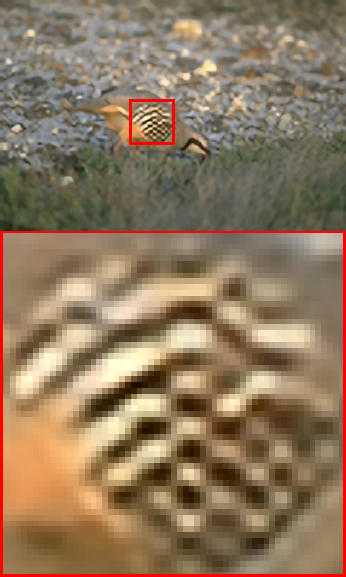}}
\subfloat[VDSR \cite{VDSR2016} \protect\\ ($29.54$, $0.8651$)]{\label{sub:VDSR:8023}\includegraphics[%
	width=0.125\textwidth, keepaspectratio]{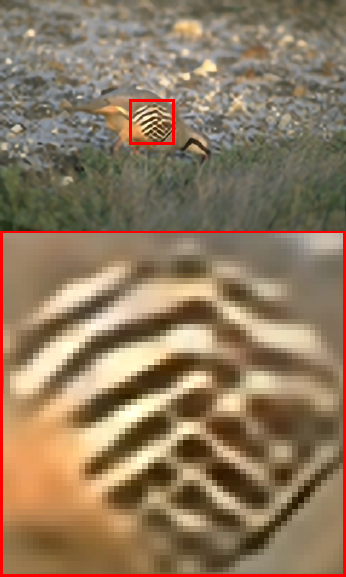}}
\subfloat[DRCN \cite{DRCN2016} \protect\\ ($\textcolor{blue}{30.29}$, $0.8653$)]{\label{sub:DRCN:8023}\includegraphics[%
	width=0.125\textwidth, keepaspectratio]{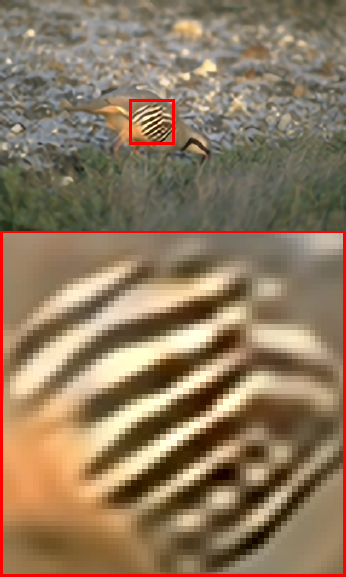}}
\subfloat[DRRN \cite{DRRN2017} \protect\\ ($29.74$, $0.8671$)]{\label{sub:DRRN:8023}\includegraphics[%
	width=0.125\textwidth, keepaspectratio]{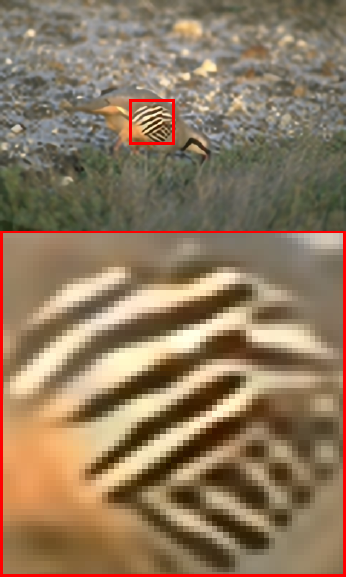}}
\subfloat[MemNet \cite{MemNet2017} \protect\\ ($30.19$, $\textcolor{blue}{0.8698}$)]{\label{sub:MemNet:8023}\includegraphics[%
	width=0.125\textwidth, keepaspectratio]{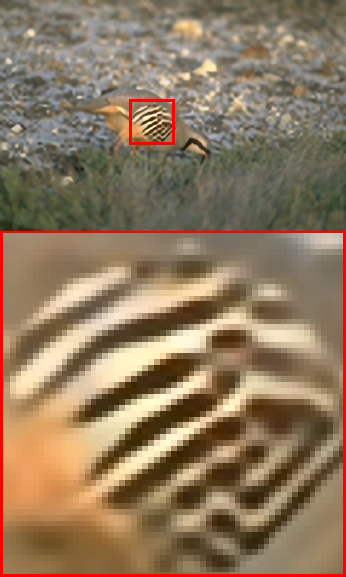}}
\subfloat[RL-CSC \protect\\ ($\textcolor{red}{30.54}$, $\textcolor{red}{0.8705}$)]{\label{sub:RL-CSC:8023}\includegraphics[%
	width=0.125\textwidth, keepaspectratio]{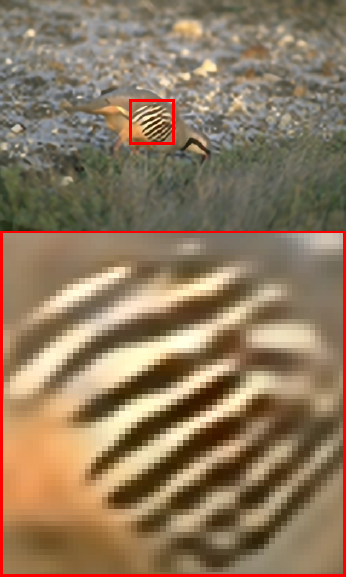}}
\vspace*{-.1cm}
\caption{\label{fig:bsd}SR results of ``8023'' from \textbf{BSD100} with scale factor $\times 4$. The direction of the stripes on the feathers is correctly restored in RL-CSC, while other methods fail to recover the pattern.}

\subfloat[Ground Truth \protect\\ (PSNR, SSIM)]{\label{sub:GT:ppt3}\includegraphics[%
	width=0.125\textwidth, keepaspectratio]{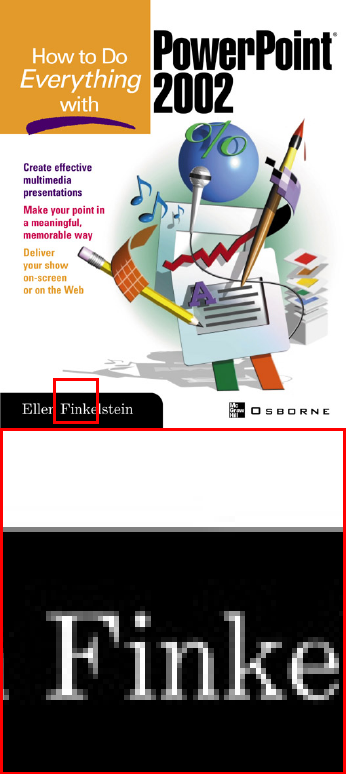}}
\subfloat[Bicubic\protect\\ ($23.71$, $0.8746$)]{\label{sub:Bicubic:ppt3}\includegraphics[%
	width=0.125\textwidth, keepaspectratio]{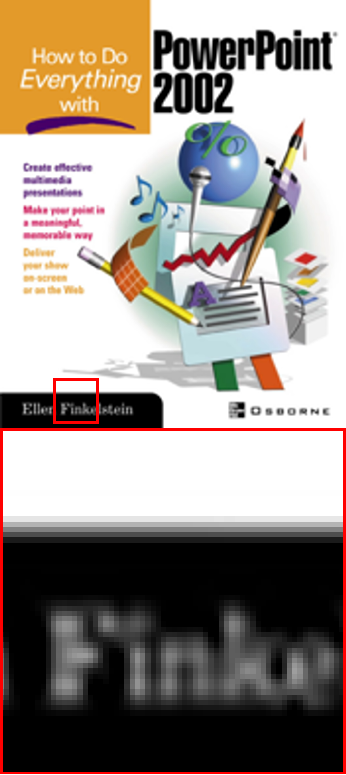}}
\subfloat[SRCNN \cite{SRCNN2016} \protect\\ ($27.04$, $0.9392$)]{\label{sub:SRCNN:ppt3}\includegraphics[%
	width=0.125\textwidth, keepaspectratio]{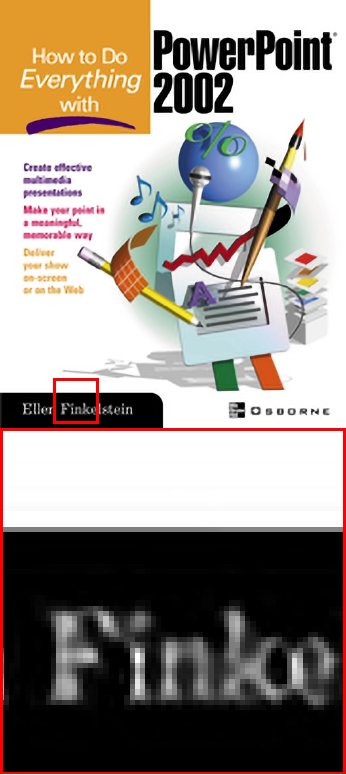}}
\subfloat[VDSR \cite{VDSR2016} \protect\\ ($27.86$, $0.9616$)]{\label{sub:VDSR:ppt3}\includegraphics[%
	width=0.125\textwidth, keepaspectratio]{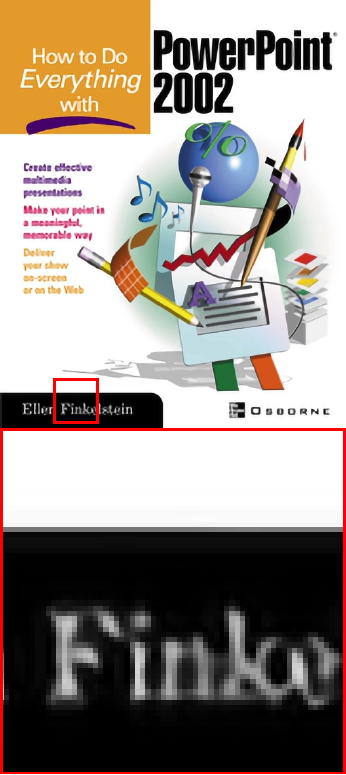}}
\subfloat[DRCN \cite{DRCN2016} \protect\\ ($27.67$, $0.9609$)]{\label{sub:DRCN:ppt3}\includegraphics[%
	width=0.125\textwidth, keepaspectratio]{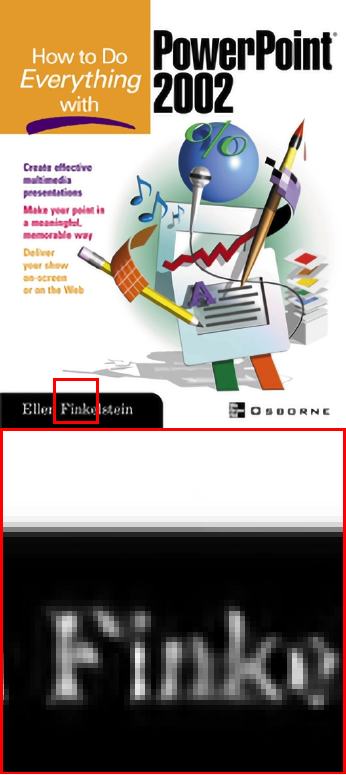}}
\subfloat[DRRN \cite{DRRN2017} \protect\\ ($\textcolor{blue}{28.70}$, $0.9702$)]{\label{sub:DRRN:ppt3}\includegraphics[%
	width=0.125\textwidth, keepaspectratio]{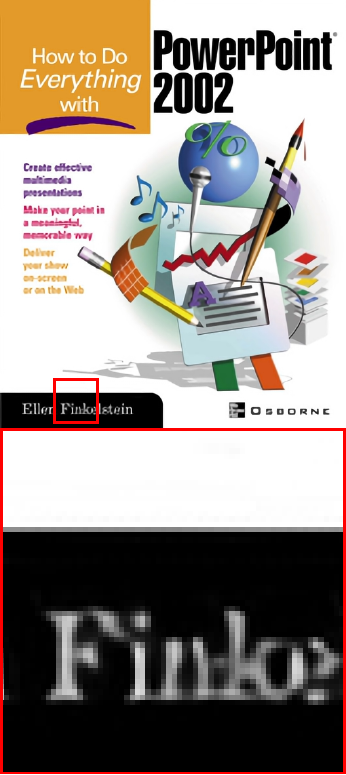}}
\subfloat[MemNet \cite{MemNet2017} \protect\\ ($\textcolor{red}{28.92}$, $\textcolor{red}{0.9711}$)]{\label{sub:MemNet:ppt3}\includegraphics[%
	width=0.125\textwidth, keepaspectratio]{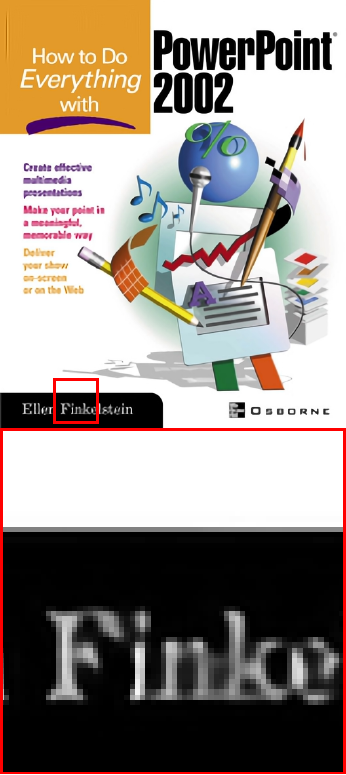}}
\subfloat[RL-CSC \protect\\ ($28.60$, $\textcolor{blue}{0.9705}$)]{\label{sub:RL-CSC:ppt3}\includegraphics[%
	width=0.125\textwidth, keepaspectratio]{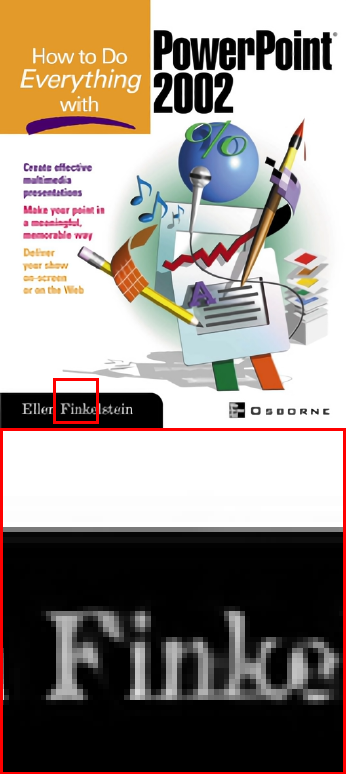}}
\vspace*{-.1cm}
\caption{\label{fig:ppt3}SR results of ``ppt3'' from \textbf{Set14} with scale factor $\times 3$. Texts in RL-CSC are sharp while character edges are blurry in other methods.}

\subfloat[Ground Truth \protect\\ (PSNR, SSIM)]{\label{sub:GT:img_076}\includegraphics[%
	width=0.125\textwidth, keepaspectratio]{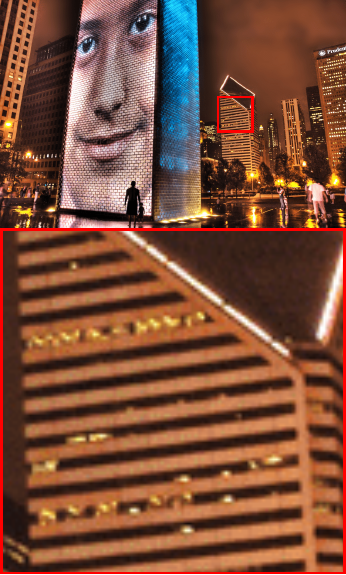}}
\subfloat[Bicubic\protect\\ ($21.57$, $0.6283$)]{\label{sub:Bicubic:img_076}\includegraphics[%
	width=0.125\textwidth, keepaspectratio]{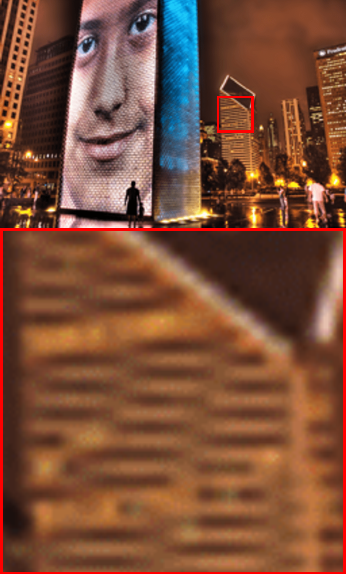}}
\subfloat[SRCNN \cite{SRCNN2016} \protect\\ ($22.03$, $0.6779$)]{\label{sub:SRCNN:img_076}\includegraphics[%
	width=0.125\textwidth, keepaspectratio]{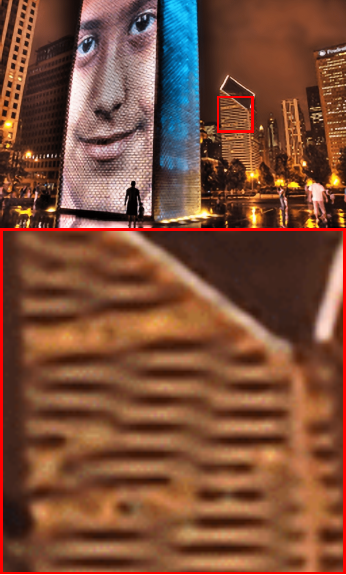}}
\subfloat[VDSR \cite{VDSR2016} \protect\\ ($22.15$, $0.6920$)]{\label{sub:VDSR:img_076}\includegraphics[%
	width=0.125\textwidth, keepaspectratio]{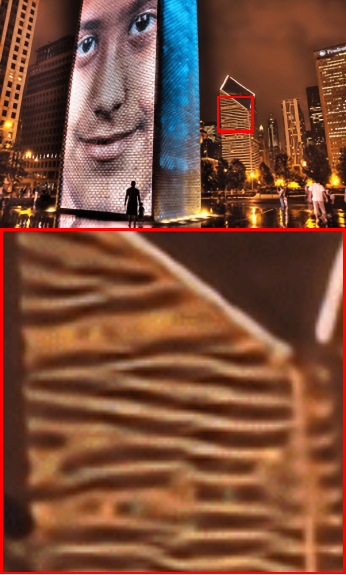}}
\subfloat[DRCN \cite{DRCN2016} \protect\\ ($\textcolor{blue}{22.11}$, $0.6867$)]{\label{sub:DRCN:img_076}\includegraphics[%
	width=0.125\textwidth, keepaspectratio]{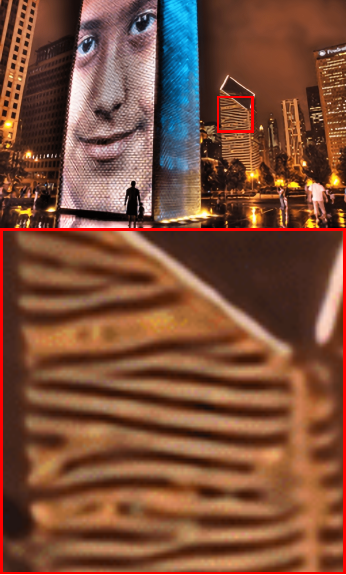}}
\subfloat[DRRN \cite{DRRN2017} \protect\\ ($21.92$, $0.6897$)]{\label{sub:DRRN:img_076}\includegraphics[%
	width=0.125\textwidth, keepaspectratio]{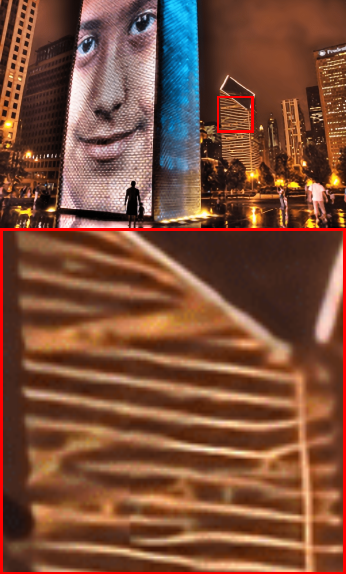}}
\subfloat[MemNet \cite{MemNet2017} \protect\\ ($22.11$, $\textcolor{blue}{0.6927}$)]{\label{sub:MemNet:img_076}\includegraphics[%
	width=0.125\textwidth, keepaspectratio]{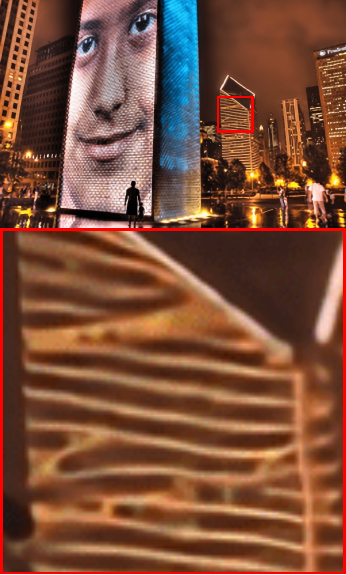}}
\subfloat[RL-CSC \protect\\ ($\textcolor{red}{22.24}$, $\textcolor{red}{0.6976}$)]{\label{sub:RL-CSC:img_076}\includegraphics[%
	width=0.125\textwidth, keepaspectratio]{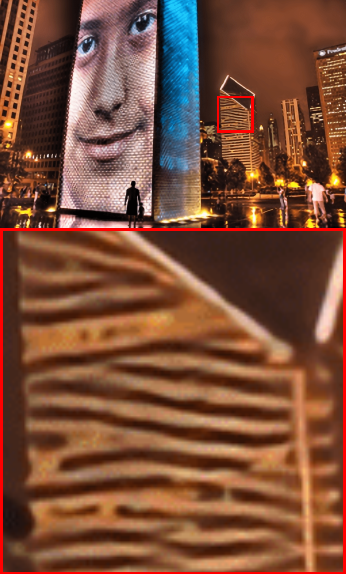}}
\vspace*{-.1cm}
\caption{\label{fig:urban}SR results of ``img076'' from \textbf{Urban} with scale factor $\times 4$. More details are recovered by RL-CSC, while others produce blurry visual results.}
\end{figure*}

Qualitative results are provided in Figures~\ref{fig:bsd}, \ref{fig:ppt3} and \ref{fig:urban}. Our method tends to produce shaper edges and more correct textures, while other images may be blurred or distorted.

\subsection{$K$, Residual Learning, Number of Filters and Batch Normalization}
\label{subsec:recursive}

\begin{figure}
\centering
\includegraphics[width=\linewidth, keepaspectratio]{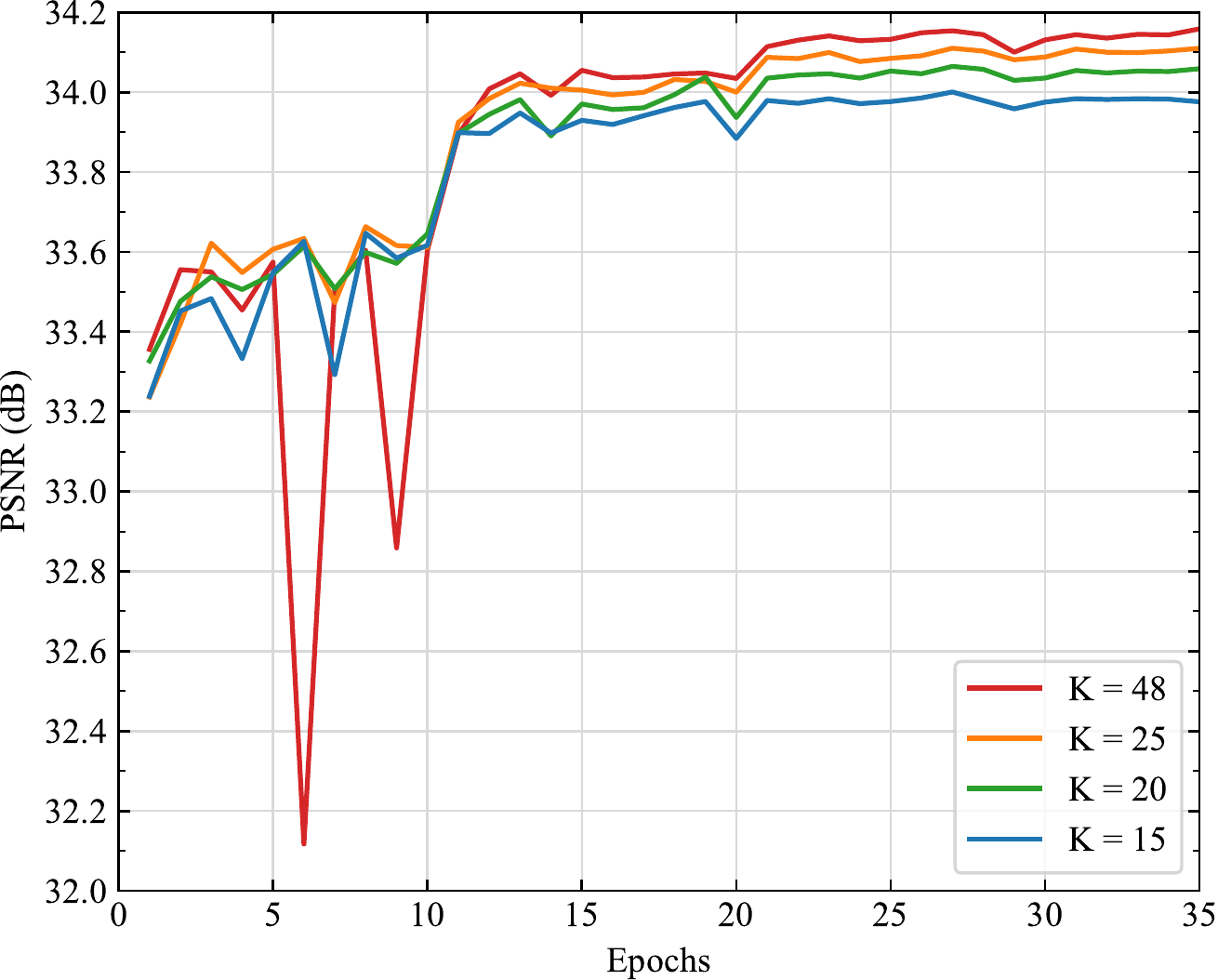}
\caption{\label{fig:psnr-k} PSNR for RL-CSC with different recursions. The models are tested under Set5 with scale factor $\times 3$.}
\end{figure}

The number of recursions $K$ is a key parameter in our model. When $K$ is increased, a deeper RL-CSC model will be constructed. We have trained and tested RL-CSC with $15$, $20$, $25$, $48$ recursions, and according to \eqref{equ:depth}, the depths of the these models are $20$, $25$, $30$, $53$ respectively. The results are presented in Figure \ref{fig:psnr-k}. The performance curves clearly show that increasing $K$ can promote the final performance ($K = 15$ $33.98$dB, $K = 20$ $34.06$dB, $K = 25$ $34.11$dB, $K = 48$ $34.16$dB), which indicates \textit{deeper is better}. Similar conclusions are observed in LISTA \cite{LISTA2010} that more iterations help prediction error decreased. However, when we extend LISTA to its convolutional version and attempt to combine the powerful learning ability of CNN, the characteristics of CNN itself must also be considered. With more recursions used, deeper networks tend to be bothered by the gradient vanishing/exploding problems. Residual learning is such a useful tool that not only solves these difficulties, but helps network converge faster. We remove the identity branch in RL-CSC and use the same parameter settings as stated in Section~\ref{subsec:details} to train the new network. The results are summarized in Table~\ref{tab:no-residual}. Without residual learning, the new network cannot even converge. We stop the training process at the $14^{th}$ epoch in advance.

{
\begin{table}
	\centering
	\begin{tabular}{|c||c|c|c|c|}
		\hline
		Epoch & $1$ & $5$ & $10$ & $13$ \\\hline
		With Residual & $33.24$ &  $33.61$ & $33.61$ & $34.02$ \\
		No Residual & $6.54$ & $6.54$ & $6.54$ & $6.54$\\\hline
	\end{tabular}
	\caption{\label{tab:no-residual} PSNR (dB) for RL-CSC and its non-residual counterpart. Tests on Set5 with scale factor $\times 3$.}
\end{table}
}

\begin{figure}
	\centering
	\includegraphics[width=\linewidth, keepaspectratio]{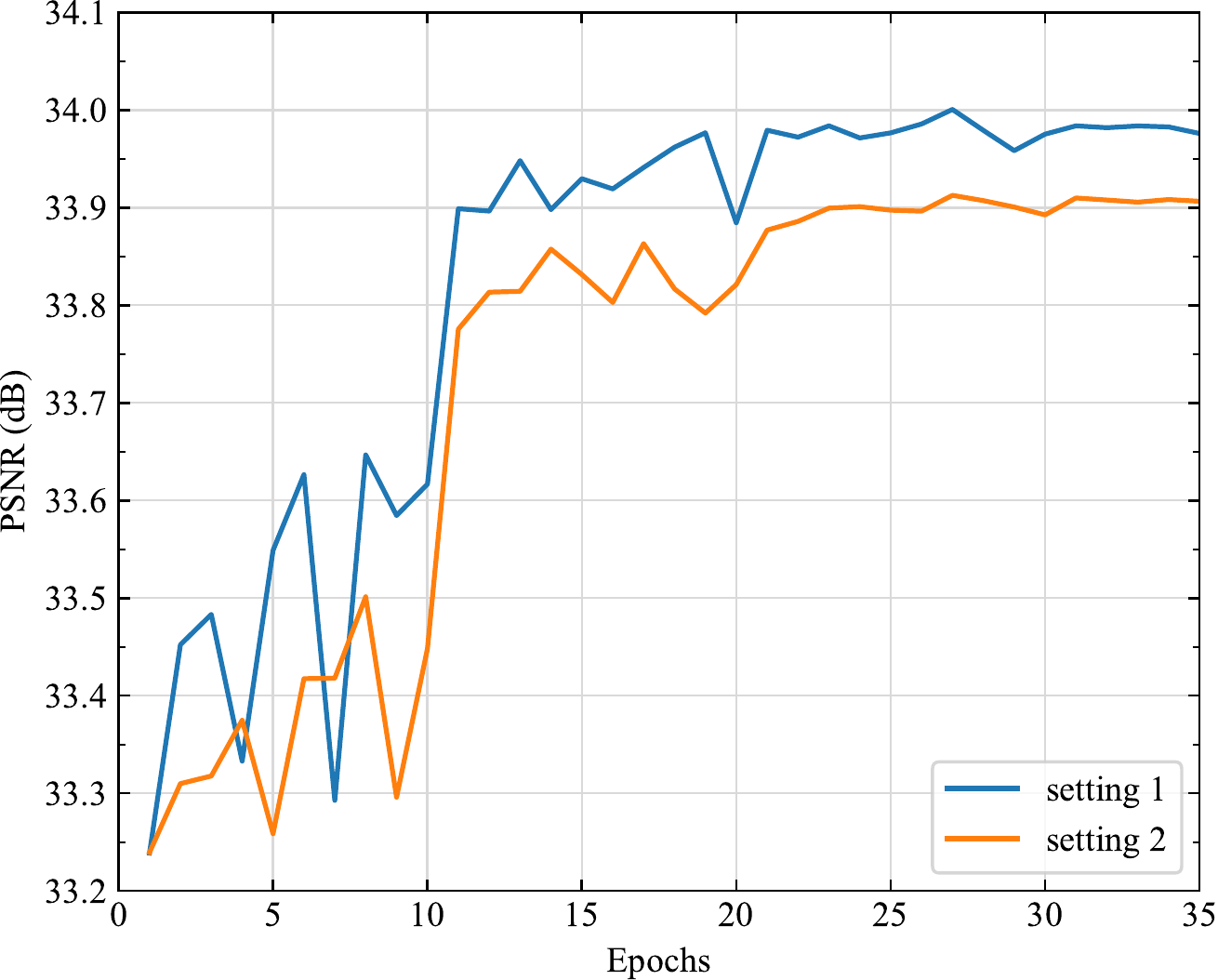}
	\caption{\label{fig:channels} Results on two types of parameter settings. The tests are conducted on Set5 with scale factor $\times 3$.}
\end{figure}

We also evaluate our model with different number of filters. Specifically, two types of parameter settings are applied:
\begin{enumerate}[i)]
	\item $\bm{F}_0\in\mathbb{R}^{128\times 1\times 3\times 3}$, $\bm{F}_1\in\mathbb{R}^{128\times 128\times 3\times 3}$, $\bm{W}_1\in\mathbb{R}^{256\times 128 \times 3\times 3}$, $\bm{S}\in\mathbb{R}^{256\times 256\times 3\times 3}$, $\bm{W}_2\in\mathbb{R}^{256\times 128\times 3\times 3}$, $\bm{H}\in\mathbb{R}^{1\times 128\times 3\times 3}$, and  $K = 15$. The total number of parameters is about $1,329$k;
	\item $\bm{F}_0\in\mathbb{R}^{128\times 1\times 3\times 3}$, $\bm{F}_1\in\mathbb{R}^{128\times 128\times 3\times 3}$, $\bm{W}_1\in\mathbb{R}^{128\times 128 \times 3\times 3}$, $\bm{S}\in\mathbb{R}^{128\times 128\times 3\times 3}$, $\bm{W}_2\in\mathbb{R}^{128\times 128\times 3\times 3}$, $\bm{H}\in\mathbb{R}^{1\times 128\times 3\times 3}$, and  $K = 15$. The total number of parameters is approximately $592$k, which is less than VDSR’s $664$k.
\end{enumerate}
Results are shown in Figure~\ref{fig:channels}. Increasing the the number of filters can benefit the performance, and our model with less parameters, \eg, $592$k, still outperforms VDSR, whose PSNR is $33.66$dB for scale factor $\times 3$ on Set5. Our final model uses the parameter settings illustrated in Section~\ref{subsec:details}. 

{
	\begin{table}
		\footnotesize
		\centering
		\begin{tabular}{|c|c|c|c|c|c|}
			\hline
			model & framework & \specialcell[c]{batch \\ size} & \specialcell[c]{patch \\ size} & memory \\\hline\hline
			DRRN ($52$) & Caffe & $128$ & $31$ & $> 12$ GB \\
			DRRN ($20$) & Caffe & $64$ & $31$ & $9,043$ MB \\
			RL-CSC ($30$) & PyTorch & $64$ & $31$ & $4,955$ MB \\
			RL-CSC ($30$) & PyTorch & $128$ & $31$ & $9,263$ MB\\
			RL-CSC ($30$) & PyTorch  & $128$ & $33$ & $9,421$ MB \\\hline
		\end{tabular}
		\caption{\label{tab:memory} GPU memory usage of different models. RL-CSC with $30$ layers are evaluated, compared to DRRN \cite{DRRN2017} with $20$ and $52$ layers. A Titan Xp with $12$ GB is used.}
	\end{table}
}

Although RL-CSC has more parameters than DRRN \cite{DRRN2017}, in our experiment we find DRRN consumes much more GPU memory resources. We test both models with different batch sizes and patch sizes of training data, and the results are summarized in Table~\ref{tab:memory}. The memory usage datas are derived from the nvidia-smi tool. Patch size of $31$ is the default setting in DRRN. Training DRRN\footnote{code can be found: \url{https://github.com/tyshiwo/DRRN_CVPR17}} of $52$ layers is difficult with one Titan Xp GPU, using the default settings given by the authors, because of the Out-of-Memory (OOM) issue. The reason is that the recursive unit of DRRN is based on the residual unit of ResNet \cite{ResNet2016}, so the BN layers are exploited, which tend to be memory intensive and increase computational burden. Guided by the analyses presented in Section~\ref{subsec:csc}, BN layers are not needed in our design. As for inference time, RL-CSC takes $0.15$ second to process a $288\times 288$ image on a Titan Xp GPU.

\section{Conclusions}

In this work, we have proposed a novel network for image super-resolution task by combining the merits of Residual Learning and Convolutional Sparse Coding. Our model is derived from LISTA so it has inherently good interpretability. We extend the LISTA method to its convolutional version and build the main part of our model by strictly following the convolutional form. Furthermore, residual learning is adopted in our model, with which we are able to construct a deeper network by utilizing more recursions without introducing any new parameters. Extensive experiments show that our model achieves competitive results with state-of-the-arts and demonstrate its superiority in SR.

{\small
\bibliographystyle{ieee}
\bibliography{reference}
}

\end{document}